\documentclass[sigconf]{acmart}
\usepackage{subcaption}

\AtBeginDocument{%
  \providecommand\BibTeX{{%
    \normalfont B\kern-0.5em{\scshape i\kern-0.25em b}\kern-0.8em\TeX}}}

\setcopyright{acmcopyright}
\copyrightyear{2023}
\acmYear{2023}
\acmDOI{10.1145/3581783.3611903}

\acmConference[MM '23]{Proceedings of the 31st ACM International Conference on Multimedia}{October 29-November 3, 2023}{Ottawa, ON, Canada}

\acmBooktitle{Proceedings of the 31st ACM International Conference on Multimedia (MM '23), October 29-November 3, 2023, Ottawa, ON, Canada}

\acmPrice{15.00}

\acmISBN{979-8-4007-0108-5/23/10}

\begin{document}

\title{Orthogonal Temporal Interpolation for Zero-Shot Video Recognition}

\author{Yan Zhu}
\affiliation{%
  \institution{School of Artificial Intelligence, University of Chinese Academy of Sciences}
  \city{Beijing}
  \country{China}}
\email{zhuyan21@mails.ucas.ac.cn}

\author{Junbao Zhuo}
\authornote{Corresponding author.}
% \authornotemark[1]
\affiliation{%
  \institution{Institute of Computing Technology, Chinese Academy of Sciences}
  \city{Beijing}
  \country{China}}
\email{junbao.zhuo@vipl.ict.ac.cn}

\author{Bin Ma}
\affiliation{%
  \institution{Meituan Inc.}
  \city{Beijing}
  \country{China}}
\email{mabin04@meituan.com}

\author{Jiajia Geng}
\affiliation{%
  \institution{Meituan Inc.}
  \city{Beijing}
  \country{China}}
\email{gengjiajia02@meituan.com}

\author{Xiaoming Wei}
\affiliation{%
  \institution{Meituan Inc.}
  \city{Beijing}
  \country{China}}
\email{weixiaoming@meituan.com}

\author{Xiaolin Wei}
\affiliation{%
  \institution{Meituan Inc.}
  \city{Beijing}
  \country{China}}
\email{weixiaolin02@meituan.com}

\author{Shuhui Wang}
\affiliation{%
  \institution{Institute of Computing Technology, Chinese Academy of Sciences}
  \city{Beijing}
  \country{China}}
\email{wangshuhui@ict.ac.cn}

\begin{abstract}
Zero-shot video recognition (ZSVR) is a task that aims to recognize video categories that have not been seen during the model training process. Recently, vision-language models (VLMs) pre-trained on large-scale image-text pairs have demonstrated impressive transferability for ZSVR. To make VLMs applicable to the video domain, existing methods often use an additional temporal learning module after the image-level encoder to learn the temporal relationships among video frames. Unfortunately, for video from unseen categories, 
we observe an abnormal phenomenon where the model that uses spatial-temporal feature performs much worse than the model that removes temporal learning module and uses only spatial feature. We conjecture that improper temporal modeling on video disrupts the spatial feature of the video. To verify our hypothesis, we propose Feature Factorization to retain the orthogonal temporal feature of the video and use interpolation to construct refined spatial-temporal feature. 
The model using appropriately refined spatial-temporal feature performs better than the one using only spatial feature, which verifies the effectiveness of the orthogonal temporal feature for the ZSVR task. 
Therefore, an Orthogonal Temporal Interpolation module is designed to learn a better refined spatial-temporal video feature during training. Additionally, a Matching Loss is introduced to improve the quality of the orthogonal temporal feature. 
We propose a model called OTI for ZSVR by employing orthogonal temporal interpolation and the matching loss based on VLMs. The ZSVR accuracies on popular video datasets (i.e., Kinetics-600, UCF101 and HMDB51) show that OTI outperforms the previous state-of-the-art method by a clear margin.
Our codes are publicly available at \href{https://github.com/sweetorangezhuyan/mm2023_oti}{\textcolor{blue}{https://github.com/yanzhu/mm2023\_oti}}.

\end{abstract}

\begin{CCSXML}
<ccs2012>
   <concept>
       <concept_id>10010147.10010178.10010224.10010225.10010228</concept_id>
       <concept_desc>Computing methodologies~Activity recognition and understanding</concept_desc>
       <concept_significance>500</concept_significance>
       </concept>
   <concept>
       <concept_id>10010147.10010178.10010187.10010193</concept_id>
       <concept_desc>Computing methodologies~Temporal reasoning</concept_desc>
       <concept_significance>500</concept_significance>
       </concept>
   <concept>
       <concept_id>10010147.10010257.10010258.10010262.10010277</concept_id>
       <concept_desc>Computing methodologies~Transfer learning</concept_desc>
       <concept_significance>500</concept_significance>
       </concept>
 </ccs2012>
\end{CCSXML}

\ccsdesc[500]{Computing methodologies~Activity recognition and understanding}
\ccsdesc[500]{Computing methodologies~Temporal reasoning}
\ccsdesc[500]{Computing methodologies~Transfer learning}

\keywords{Zero-shot Video Recognition, Visual Language Models,
Temporal Modeling, Transfer Learning.}

\maketitle

\section{Introduction}

\begin{figure*}
	\centering
        \begin{minipage}[]{0.3\textwidth}
		\includegraphics[width=\textwidth]{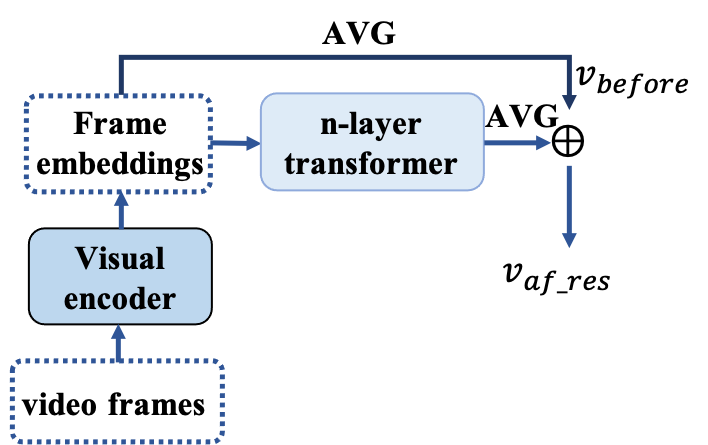}
            \label{fg:text4vis}
		\subcaption{Details of the video branch of Text4Vis~\cite{wu2022transferring}.}	
	\end{minipage} 
 \quad
	\begin{minipage}[]{0.305\textwidth}
		\includegraphics[width=\textwidth]{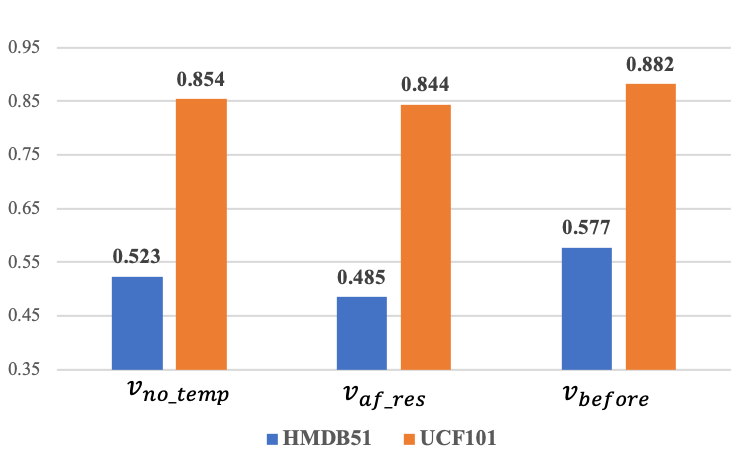}
            \label{fg:tm_text4vis}
		\subcaption{ZSVR accuracies of pre-trained Text4Vis with the backbone VIT-L/14.}	
	\end{minipage} 
        \quad
	\begin{minipage}[]{0.32\textwidth}
		\includegraphics[width=\textwidth]{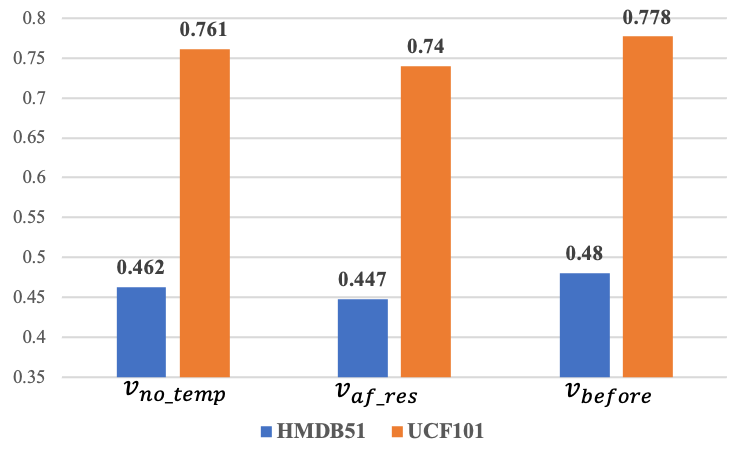}
            \label{fg:tm_bs}
		\subcaption{ ZSVR accuracies of the baseline model (see Section~\ref{sec:bsmodel}) with the backbone VIT-B/32.}	
	\end{minipage}
 % \vspace{-1ex}
	\caption{The abnormal phenomenon occurs on zero-shot video recognition task based on visual language model.}
	\label{fig:problem}
\end{figure*}

As video gradually becomes the mainstream content delivery form in multimedia, video recognition~\cite{vr} task has a wide range of applications, providing us with more convenient and vivid experience in life. Training a video recognition model with high performance usually requires a large amount of annotated video. However, collecting labeled data for a huge number of video categories is impractical as annotating videos requires much more effort compared with image annotation~\cite{ts-gcn,tsrl}. To overcome these issues, Zero-Shot Video Recognition (ZSVR) task has attracted wide attention. The goal of ZSVR is to enable the model to identify categories that have not appeared during the training process, by leveraging video content understanding and semantic relations among categories~\cite{e2e, gao2020ci,ts-gcn,tsrl,ER,zhang2018visual}.

Recently, the visual language models (VLMs)\cite{clip,jia2021scaling,yu2022coca} provide extraordinary video content understanding ability and semantic knowledge. VLMs are trained with large-scale image-text pairs using contrastive learning effectively align the visual space and the semantic space. CLIP\cite{clip}, which is an image-based VLM, is highly flexible and general. It enables a wide range of vision tasks to be performed in a zero-shot manner. 
However, CLIP, which is built on the image domain, is not directly applicable to video. To make CLIP suitable for the video domain, several methods such as X-CLIP\cite{x-clip}, Text4Vis\cite{wu2022transferring}, ViFi-CLIP\cite{hanoonavificlip} and BIKE\cite{bike} have been proposed recently. For ZSVR, temporal information is unique to video compared to static images. Therefore, it is necessary to employ some strategies that enable the model to learn temporal feature for video. Using an additional trainable subnetwork, such as an n-layer transformer, is a simple way to bridge the modality gap between images and videos\cite{wu2022transferring,vtn,vivit,avt,hanoonavificlip}. However, it is uncertain whether the temporal relationships learned from videos of seen categories can be effectively transferred to videos of unseen categories~\cite{9882310,Zhuo_2019_CVPR}.

To answer the above question, we conduct some experiments based on Text4Vis\cite{wu2022transferring}. Text4Vis trains a temporal learning module consisting of six-layer transformer to learn the temporal relationship among video frames, as shown in Figure~\ref{fig:problem}(a). 
In Text4Vis, feature \textbf{$v_{af\_res}$} contains both spatial and temporal features, while feature \textbf{$v_{before}$} only contains spatial feature. We find that the model using the feature \textbf{$v_{af\_res}$} performs significantly worse than the one using feature \textbf{$v_{before}$}, as shown in Figure~\ref{fig:problem}(b), for both HMDB51 (\textbf{\textcolor{blue}{blue}}) and UCF101 (\textbf{\textcolor{orange}{orange}}). 
To investigate whether the temporal learning module leads to the performance degradation, we train a model without temporal learning that only learns the spatial feature \textbf{$v_{no\_temp}$}. Our results in Figure~\ref{fig:problem}(b) show that the model using the feature \textbf{$v_{no\_temp}$} performs better than the one using feature \textbf{$v_{af\_res}$}, but worse than the one using feature $v_{before}$.
Furthermore, as shown in Figure~\ref{fig:problem}(c), this \textbf{abnormal phenomenon} is quite common and also occurs in our constructed baseline, where the six-layer transformer in Text4Vis is replaced with a one-layer transformer.

We conjecture that the abnormal phenomenon observed for the ZSVR task is caused by improper temporal modeling of video from unseen categories, which disrupts the spatial feature $v_{before}$. To verify our hypothesis, we eliminate the spatial feature in feature $v_{af\_res}$ to better understand the effect of the temporal feature on the spatial feature $v_{before}$. 
To achieve this, we propose a method called \textbf{Feature Factorization}. Specifically, the temporal feature decomposed from feature $v_{af\_res}$ is orthogonal to the feature $v_{before}$. We then fuse feature $v_{before}$ with the orthogonal temporal feature using different interpolation weights to obtain a refined spatial-temporal feature. We conduct experiments using the refined feature to explore the impact of the temporal feature. Our experimental results (see Section~\ref{sec:fea_fac}) confirm that appropriate fusion is effective in enhancing the feature representation ability.

Based on the effectiveness of orthogonal temporal feature, we propose an \textbf{Orthogonal Temporal Interpolation} module. This module also inspired by~\cite{yang2021dolg}, where local feature and global feature (like the spatial-temporal feature $v_{af\_res}$ and the spatial feature $v_{before}$) can be mutually reinforced to produce the final representative feature. The module is used during training to learn a better refined spatial-temporal feature for video.
We use cross-entropy loss to maximize the similarity between the refined spatial-temporal feature and the semantic feature of category that matches with video.
However, we still observe a performance gap between the model using feature $v_{af\_res}$ and the one using feature $v_{before}$. This gap limits the quality of the learned orthogonal temporal feature and further restricts the quality of the refined spatial-temporal feature. To address this issue, we introduce the \textbf{Matching Loss} to encourage better learning of orthogonal temporal feature for video.

Leveraging the proposed \textbf{O}rthogonal \textbf{T}emporal \textbf{I}nterpolation and Matching Loss, we propose a novel framework for ZSVR based on CLIP, called \textbf{OTI}. We carry out extensive experiments on public datasets (i.e., HMDB51, UCF101 and Kinetics-600), to validate the effectiveness of the proposed techniques. Experimental results show the proposed OTI achieves new state-of-the-art performances.

\section{Related Work}
\subsection{Aligning Video and Category}
There are roughly three main kinds of approaches for aligning video and category in zero-shot video recognition (ZSVR) task. The first kind of approaches are indirect methods that represent video using visual attributes~\cite{ts-gcn,tsrl,ER,gao2020learning}. For instance, Liu et al.~\cite{liu2011recognizing} represent video using a set of visual attributes and then align those attributes with categories for achieving the ZSVR task. TS-GCN~\cite{ts-gcn} extracts shared attributes from videos of the seen categories, while TSRL~\cite{tsrl} constructs shared attributes from web images and also considers unseen categories. ER~\cite{ER} extends video knowledge by generating elaborative concepts of the attributes using WordNet and constructs elaborative descriptions sentences from action class names using Wikipedia and Dictionary. However, a common issue with these methods is that they tend to lose the temporal information..

The second kind of approaches involve directly extracting high-level visual embeddings from videos using deep learning techniques, which has become possible with recent advancements in the field. Methods such as ~\cite{lin2019tsm,ER,rest} attempt to extract visual embeddings directly from videos. A simple method is to use 3D CNNs, but this approach results in high computational costs. To address this issue, improved models such as I3D~\cite{i3d} and TSM~\cite{lin2019tsm} have been adopted in subsequent works. For categorical text, the semantic embedding is computed using Word2Vec~\cite{mikolov2013efficient}. The goal of this kind of approach is to align the visual and semantic embeddings in the feature space.

The third kind of approaches involve using vision language models (VLMs) that leverage large-scale image-text pairs for training and facilitate effective alignment between visual and semantic spaces. Paired visual encoder and textual encoder from VLMs are used to align video and category, as demonstrated in works such as ~\cite{wu2022transferring,bike,hanoonavificlip,CLIP-Adapter,PointCLIP,Tip-Adapter,VideoCLIP,x-clip}. For example, X-CLIP~\cite{x-clip} trains the parameters of the visual encoder and textual encoder from CLIP~\cite{clip}, while Text4Vis~\cite{wu2022transferring} freezes the textual encoder of CLIP, optimizes the visual encoder of CLIP and a six-layer transformer using cross-entropy loss. BIKE~\cite{bike} generates textual auxiliary attributes for each video to complement video recognition. 
These works have shown excellent performance on the ZSVR task, demonstrating that VLMs have a strong ability to align video and category.

\subsection{Temporal Information Learning}

Compared with the image domain, the video domain provides additional temporal information that is crucial for accurately classifying videos. Convolutional networks have been the standard backbone architecture for temporal learning for a long time. To achieve effective temporal modeling, researchers have developed plug-and-play temporal modules for 2D CNN backbones, such as those proposed in~\cite{li2020tea,liu2020teinet,liu2021tam,tran2018closer,qiu2017learning}.
Models based on self-attention, such as VTN~\cite{vtn}, ViViT~\cite{vivit}, AVT~\cite{avt}, ViFiCLIP~\cite{hanoonavificlip}, and Text4Vis~\cite{wu2022transferring}, learn additional parameters to capture temporal features in videos. Late fusion is a simple and popular approach where a temporal learning module is added after the frame-level encoder. However, late fusion encodes each frame separately, which may not fully utilize the temporal cues. To address this issue, X-CLIP~\cite{x-clip} replaces spatial attention with cross-frame attention, allowing for global spatial-temporal modeling of all frames. Additionally, BIKE~\cite{bike} weights video frames based on matched semantic features of categories to capture temporal saliency.
However, these models only treat zero-shot video recognition as a subtask and do not seriously consider its characteristics.
Therefore, our focus is on investigating whether the temporal relationships learned by videos from seen categories are applicable to videos from unseen categories. 
We empirically observe that improper temporal modeling in existing methods like Text4Vis can destroy the spatial feature of video in ZSVR. Therefore, we propose orthogonal temporal interpolation module to learn a better refined spatial-temporal feature for ZSVR.

\section{Our Approach}

\begin{figure*}
 \centering
  \includegraphics[width=\textwidth]{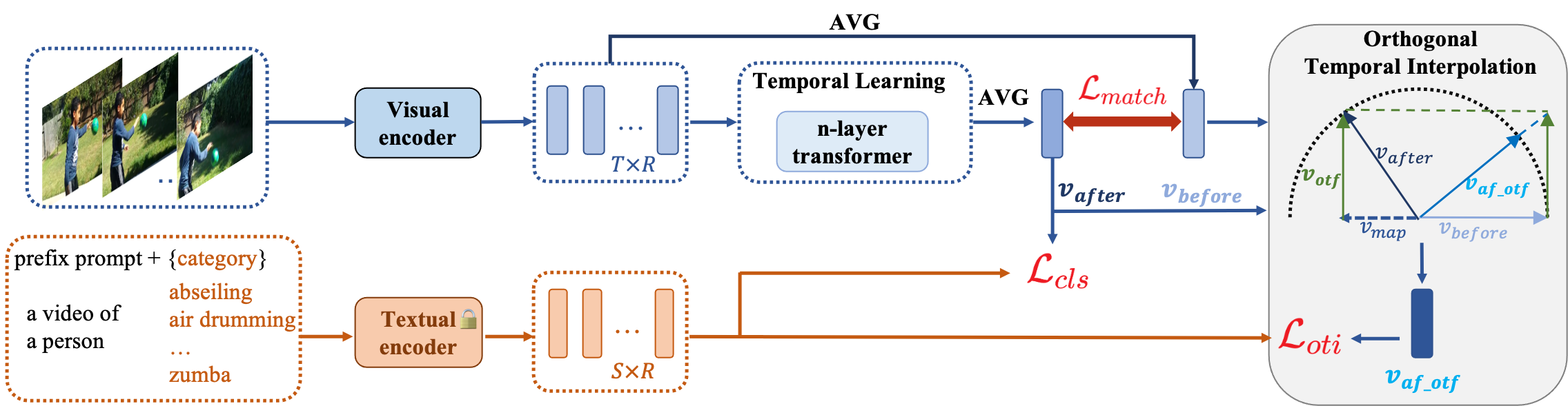}
  \caption{The framework of OTI. OTI has two branches: the video branch and the category branch. For the video branch, firstly, we use a visual encoder and AVG to map video frames to spatial feature $v_{before}$. Then we use a Temporal Learning module to obtain the spatial-temporal feature $v_{after}$. Next, we use Orthogonal Temporal Interpolation module to construct the refined spatial-temporal feature $v_{af\_otf}$. For the category branch, textual encoder is used to map description of categories into semantic features. Both the visual encoder and textual encoder come from the VLMs. We use the cross-entropy loss $\mathcal L_{cls}$ between the feature $v_{after}$ and semantic features of seen categories, the cross-entropy loss $\mathcal L_{oti}$ between the feature $v_{af\_otf}$ and semantic features of seen categories, and the matching loss $\mathcal L_{match}$ (MSE loss between $v_{before}$ and $v_{after}$) to train the whole model.}
 
  \label{fig:framework}
\end{figure*}

\subsection{Problem Setting}
We focus on the inductive zero-shot setting for the zero-shot video recognition (ZSVR) task in this paper. For this setting, unlabeled videos from unseen categories are unavailable during the training phase. The model is trained with the videos of seen categories and evaluated with those of unseen categories. For the seen categories, there are $N_S$ labeled videos $D^S=\left \{ v_i^S,y_i^S \right \}_{i=1}^{N_S}$ with $S$ different categories. Besides, there are $N_U$ videos $D^U=\left \{ v_e^U,y_e^U \right \}_{e=1}^{N_U}$ from $U$ unseen categories. It is important to note that there is no overlap between the categories of the seen and unseen, $Y_S\bigcap Y_U=\varnothing$.

\subsection{Preliminary} 
\label{sec:bsmodel}

We present a baseline model for zero-shot video recognition (ZSVR) task based on the pre-trained Contrastive Language-Image Pre-Training (CLIP)~\cite{clip}. CLIP is a popular visual-language model (VLM) that comprises two main modules: a visual encoder and a textual encoder. It is obtained through contrastive learning on a vast collection of image-text pairs and has certain zero-shot recognition capabilities.
Our baseline model consists of two branches: video branch and category branch. The video branch maps videos to the visual space, while the category branch maps categories to the semantic space. The visual space and semantic space have the same dimension $d$. The key modules of both branches are the visual encoder $f(\cdot|\theta_v)$ and the textual encoder $f(\cdot|\theta_t)$ from CLIP. The entire baseline model is trained with cross-entropy loss.

For the \textbf{video branch}, we take the video frames $\left \{ I_{i,1},I_{i,2},\dots  I_{i,T}\right \} $ as inputs for a given video $v_i^S$, where $T$ represents the number of frames. Then the video $v_i^S$ is encoded with the encoder $f(\cdot|\theta_v)$ as $V_i^S=\left \{ f\left (I_{i,1} \right |\theta_v), f\left (I_{i,2} \right |\theta_v),\\ \dots, f\left (I_{i,T} \right |\theta_v)\right \} $. 
We average the features $V^S$ (we drop the subscript i for convenience) along the frame dimension to obtain feature $v_{before}$, which only contains the spatial feature. 
To capture the temporal relationships among video frames, we introduce a temporal learning module $F(\cdot|\theta_{temp})$, which is composed of an n-layer transformer. This module takes $V^S$ as input and produces the output feature $v_{after}\in R^d$ as follows:
\begin{equation}
\label{eq:after}
  v_{after}=AVG(F(V^S|\theta_{temp})).
\end{equation}
The temporal learning module learns the temporal dynamics of video frames, which enables the model to capture significant temporal cues in the video stream. Therefore, the feature $v_{after}$ contains both the spatial and temporal features. 
Previous studies~\cite{resnet, wu2022transferring} have shown that residual connection can prevent information loss and alleviate overfitting. So residual connection is adopted to obtain an enhanced spatial-temporal feature $v_{af\_res}$ for video as follows:
\begin{equation}
v_{af\_res}=AVG(F(V^S|\theta_{temp})+V^S)=v_{after}+v_{before}.
  \label{eq:af_res}
\end{equation}

For the \textbf{category branch}, to generate informative description for category, we use the prefix prompt ``a video of a person'', which is a common practices in previous works~\cite{wu2022transferring,bike}, and append the category name to it. The description of category $c$ is then encoded with the encoder $f(\cdot|\theta_t)$ to obtain the semantic feature $c_{cls}$, as
\begin{equation}
  c_{cls}=f(c|\theta_t).
\end{equation}

We train the baseline model with the cross-entropy loss $\mathcal L_{cls}'$, which is formulated as follows:
\begin{equation}
\mathcal L_{cls}'=-\frac{1}{N_S}\sum_{i=1}^{N_S}\sum_{k=1}^{S}y_{i,k}^Slog(\frac{exp(CS({v_{af\_res}^i},c_{cls}^k)) }{\sum_{j=1}^{S} exp(CS({v_{af\_res}^i},c_{cls}^j))} ).
\label{eq:loss1}
\end{equation}
where $y_{i,k}^S$ and $CS({v_{after}^i},c_{cls}^k)$ are the ground-truth label and the cosine similarity score for the $i_{th}$ training video from $k_{th}$ seen category. 
During the training phase, the parameters $\theta_v$ and $\theta_t$ of the visual and textual encoders are initialized with weights from a pre-trained CLIP. The parameters $\theta_{temp}$ of temporal learning module are initialized randomly. We freeze the parameters $\theta_t$ and train the parameters $\theta_v$ and $\theta_{temp}$ with the loss $\mathcal L_{cls}'$.

For the inference of a video from unseen categories, we extract its enhanced spatial-temporal feature $v_{af\_res}$ and semantic features of all unseen categories through the video branch and category branch respectively. Then, we calculate the cosine similarity scores between feature $v_{af\_res}$ and semantic features of all unseen categories. Finally, we select the category with the highest similarity score as the predicted category of this video.

\subsection{Feature Factorization}
\label{sec:fea_fac}
\begin{figure}[!h]
  \centering
  \includegraphics[width=0.35\linewidth]{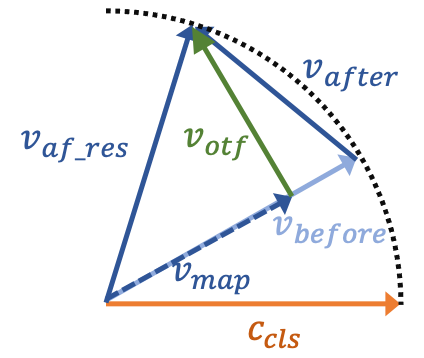}
  \caption{The features relationships obtained based on ZSVR accuracies of our baseline model trained with $\mathcal L_{cls}'$.}
  \label{fig:temporal fac}
\end{figure}
\begin{figure}
        \begin{minipage}[]{0.22\textwidth}      
		\includegraphics[width=\textwidth]{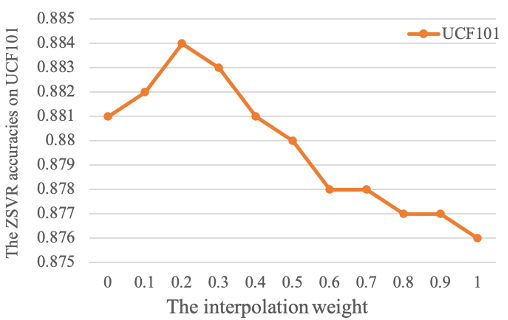}
		\subcaption{The trend of ZSVR accuracies of pre-tained TextVis with the backbone VIT-L/14. }
	\end{minipage} 
        \quad
	\begin{minipage}[]{0.22\textwidth}
		\includegraphics[width=\textwidth]{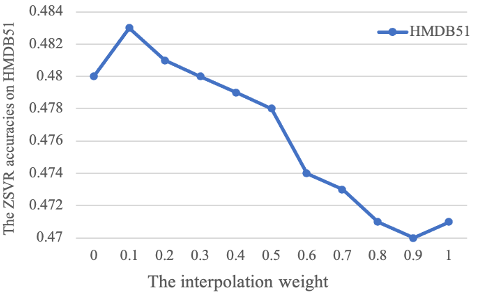}
		\subcaption{The trend of ZSVR accuracies of our baseline model with the backbone VIT-B/32. }	
	\end{minipage} 
	\caption{The trends of ZSVR accuracies of the model using the feature $v_{af\_otf}$ with various $\lambda$. }
	\label{fig:conjecture}
\end{figure}

For the ZSVR task, we observe an abnormal phenomenon where the model using the enhanced spatial-temporal feature $v_{af\_res}$ performs worse than the one using only the spatial feature $v_{before}$.
We conjecture that the improper temporal modeling for the video disrupts the spatial feature.
To verify our conjecture, we propose a method called \textbf{feature factorization}. This method first eliminates the mixed spatial feature $v_{map}$ in $v_{af\_res}$ and retains the temporal feature $v_{otf}$. Specifically, feature $v_{otf}$ is orthogonal to the feature $v_{before}$. We then fuse feature $v_{before}$ and feature $v_{otf}$ using different interpolation weights to construct a refined spatial-temporal feature $v_{af\_otf}$. Finally, we use feature $v_{af\_otf}$ to explore the impact of the temporal feature on the spatial feature.
The symbols and the relationships between features are shown in Figure~\ref{fig:temporal fac}. 
We use symbol $c_{cls}$ to represent the ideal semantic feature of category that matches with video.
\textbf{Ideally, the closer the video feature is to feature $c_{cls}$, the higher ZSVR accuracy obtained.}
We normalize the features following Text4Vis\cite{wu2022transferring}, such that $\left \| v_{before}  \right \|=\left \| v_{af\_res}  \right \|=\left \| c_{cls}  \right \|= 1$, where $\left \| \cdot  \right \|$ represents the magnitude of the feature. 
Therefore, the features can be represented as vectors with their end points on a circle with unit length.
A higher ZSVR accuracy indicates that the video feature is closer to the feature $c_{cls}$, which means that the angle between them is smaller. Based on the above abnormal phenomenon, where ZSVR accuracy of the model using feature $v_{af\_res}$ is worse than of the model using feature $v_{before}$, we can conclude that the angle between feature $v_{af\_res}$ and feature $c_{cls}$ is bigger than the angle between $v_{before}$ and $c_{cls}$.
To elaborate on the first step, we begin by obtaining the projection feature $v_{map}$ of feature $v_{af\_res}$ onto feature $v_{before}$. The feature $v_{map}$ can be formulated as:
\begin{equation}
     v_{map}=\frac{\left \langle v_{af\_res},v_{before} \right \rangle  }{\left \|  v_{before}  \right \|_2 ^2}v_{before},
    \label{eq:map}
\end{equation}
where $\left \langle \cdot,\cdot \right \rangle $ is the dot product of two features, $\left \|  v_{before}  \right \|_2$ is the $L2$ norm of feature $v_{before}$. The projection feature $v_{map}$ only contains spatial feature that is mixed in feature $v_{af\_res}$. 
Next, we can obtain the retained temporal feature $v_{otf}$ using the following formula:
\begin{equation}
    v_{otf}=v_{af\_res}-v_{map}.
    \label{eq:otf}
\end{equation}
From Figure~\ref{fig:temporal fac}, it is evident that the feature $v_{otf}$ is orthogonal to the spatial feature $v_{before}$. Additionally, 
we can see that $v_{otf}$ can also be the orthogonal feature of feature $v_{after}$ onto feature $v_{before}$.

In the second step, we construct refined spatial-temporal feature $v_{af\_otf}$ by fusing feature $v_{before}$ with the feature $v_{otf}$ using different interpolation weights. The formula is as follows:
\begin{equation}
v_{af\_otf}=v_{before}+\lambda v_{otf},
\label{eq:newf}
\end{equation}
where $\lambda \in [0,1]$ is an interpolation weight, representing the degree to which the feature $v_{otf}$ is retained.

We conduct ZSVR experiments using feature $v_{af\_otf}$ to verify our conjecture. The experimental results are presented in Table~\ref{fig:conjecture}. We replace feature $v_{otf}$ with feature $v_{after}$ to synthesize a new spatial-temporal feature for ZSVR. The comparison results are presented in Appendix~\ref{supp:otf-af}. 
It is evident that the ZSVR accuracy undergoes a slight improvement with small $\lambda$. % upon the addition of a small amount of orthogonal temporal feature
However, as the $\lambda$ increases, the ZSVR accuracy begins to continuously decline and falls below the accuracy of the model using the feature $v_{before}$. The trends of ZSVR accuracy confirm our conjecture that improper temporal modeling for video disrupts the spatial feature. Furthermore, the experiment results demonstrate that appropriate fusion is effective in enhancing the feature representation ability for the ZSVR task.

\subsection{Orthogonal Temporal Interpolation for ZSVR}
\label{sec:otis}

To better learn proper temporal feature for video, we propose a framework for ZSVR task based on CLIP~\cite{clip}, called \textbf{OTI}, as shown in Figure~\ref{fig:framework}. The key module in our framework is the proposed \textbf{Orthogonal Temporal Interpolation} module. We draw inspiration for this module from~\cite{yang2021dolg}, which shows that local and global components (like the spatial-temporal feature $v_{af\_res}$ or $v_{after}$ and the spatial feature $v_{before}$) can be mutually reinforced to produce a final representative descriptor with objective-oriented training. Therefore, we hope that this module can learn a better refined spatial-temporal feature $v_{af\_otf}$ for video during the model training process. This module includes two stages: feature generation and loss calculation. The module takes spatial feature $v_{before}$, spatial-temporal feature $v_{after}$ and semantic features of all seen categories as inputs. 
Due to the feature $v_{otf}$ is also the orthogonal feature of feature $v_{after}$ onto feature $v_{before}$ and more spatial feature is added to feature $v_{af\_res}$ compared with feature $v_{after}$. Therefore, to fully exploit the potential of the temporal learning module, we remove the residual connection, then the loss $\mathcal L_{cls}'$ in Eqn.~(\ref{eq:loss1}) is modified to $\mathcal L_{cls}$, and can be rewritten as follows:
\begin{equation}
\mathcal L_{cls}=-\frac{1}{N_S}\sum_{i=1}^{N_S}\sum_{k=1}^{S}y_{i,k}^Slog(\frac{exp(CS({v_{after}^i},c_{cls}^k)) }{\sum_{j=1}^{S} exp(CS({v_{after}^i},c_{cls}^j)) } ).
% \label{eq:loss1}
\end{equation}

For \textbf{feature generation}, in detail, the projection feature $v_{map}$, the orthogonal temporal feature $v_{otf}$ and the refined spatial-temporal feature $v_{af\_otf}$ are obtained according to the Eqn.~(\ref{eq:map}), Eqn.~(\ref{eq:otf}) and Eqn.~(\ref{eq:newf}), respectively. For \textbf{loss calculation}, to learn more effective feature $v_{af\_otf}$, we aim to ensure that it is closer to the semantic feature of category that matches with video. To achieve this, we introduce an additional loss $\mathcal L_{oti}$ to jointly train the entire model. The cross-entropy loss $\mathcal L_{oti}$ is formulated as follows:
\begin{equation}
\mathcal L_{oti}=-\frac{1}{N_S}\sum_{i=1}^{N_S}\sum_{k=1}^{S}y_{i,k}^Slog(\frac{exp(CS({v_{af\_otf}^i},c_{cls}^k)) }{\sum_{j=1}^{S} exp(CS({v_{af\_otf}^i},c_{cls}^j)) } ).
\end{equation}

However, when we train the model with loss $\mathcal L_{cls}$ and $\mathcal L_{oti}$ as described above, we still observe that ZSVR accuracy of the model using feature $v_{after}$ is worse than that of the model using feature $v_{before}$ (as shown in Table~\ref{tab:loss}). 
The gap limits the quality of the learned orthogonal temporal feature and further limits the quality of the refined spatial-temporal feature. To address this issue, we introduce an additional \textbf{Matching Loss} to encourage the whole model to learn better orthogonal temporal feature for video. Specifically, we use the Mean Squared Error (MSE) loss function to calculate the matching loss.
\begin{equation}
\mathcal L_{match}=MSE(v_{after},v_{before}). 
\end{equation}

In summary, the overall loss of our model can be expressed as:
\begin{equation}
\label{eq:all_loss}
\mathcal L=\alpha \mathcal L_{cls}+ \beta \mathcal L_{oti}+ \gamma
\mathcal L_{match}.
\end{equation}
$\alpha$, $\beta$ and $\gamma$ are the weights for the losses.
\begin{figure}[!h]
    \centering
    \includegraphics[width=0.5\linewidth]{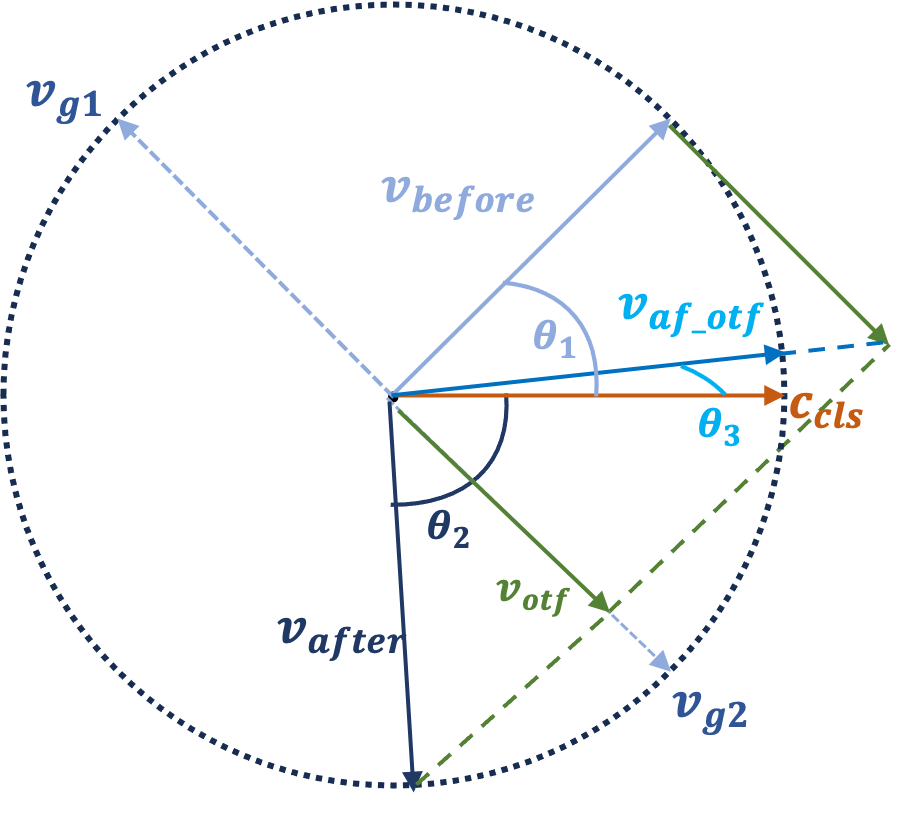}
    \caption{
    The features relationships obtained based on ZSVR accuracies of the model trained with $\mathcal L_{cls}$ and $\mathcal L_{oti}$.}
    \label{fig:match}
\end{figure}

We explain why the additional loss $\mathcal L_{match}$ is effective  in the following. 
As shown in the Figure~\ref{fig:match}, for a single video, it has spatial feature $v_{before}$, spatial-temporal feature $v_{after}$ and spatial-temporal feature $v_{af\_otf}$. We use symbol $c_{cls}$ to represent the ideal semantic feature of category that matches with video.
There are two orthogonal features of $v_{before}$, namely feature $v_{g1}$ and feature $v_{g2}$.
The orthogonal feature $v_{otf}$ of $v_{after}$ on feature $v_{before}$ falls on either feature $v_{g1}$ or feature $v_{g2}$.
The angle between $v_{before}$ and $c_{cls}$, $v_{after}$ and $c_{cls}$, $v_{af\_otf}$ and $c_{cls}$ are denoted as $\theta_1$, $\theta_2$ and $\theta_3$, respectively. Therefore, The angle between $v_{before}$ and $v_{after}$ is $\theta_1+\theta_2$. 
According the ZSVR accuracies of the model trained with $\mathcal L_{oti}$ and $\mathcal L_{cls}$ (as shown in Table~\ref{tab:loss}), we clearly know that the ZSVR accuracy is highest for the model using feature $v_{af\_otf}$, followed by the model using feature $v_{before}$, and lowest for the model using feature $v_{after}$.
Therefore, $\theta_3<\theta_1<\theta_2$, feature $v_{otf}$ falls on feature $v_{g2}$. This is why the features relationships are shown in Figure~\ref{fig:match}.  
From the figure, we can infer that $cos(\theta_1+\theta_2-90^\circ)$ can be regarded as the degree of overlap of feature $v_{otf}$ onto $v_{g2}$, so $v_{otf}$ is updated as 
$v_{otf}=cos(\theta_1+\theta_2-90^\circ)v_{g2}$. 

In brief, during the training process, $\mathcal L_{oti}$ is used to encourage the feature $v_{af\_otf}$ to be close to the feature $c_{cls}$. $\mathcal L_{match}$ is used to encourage feature $v_{after}$ to close the feature $c_{before}$. During the training process, the sum of angle $\theta_1+\theta_2$ decreases, and there are two possible scenarios that may occur:
\begin{itemize}
\item {}$\theta_1+\theta_2>90^\circ$: As the decrease in $\theta_1+\theta_2$, $cos(\theta_1+\theta_2-90^\circ)$ increases, causing $v_{otf}$ to move closer to $v_{g2}$. The angle $\theta_3$ between $v_{af\_otf}$ and $c_{cls}$ becomes smaller, and $v_{af\_otf}$ moves closer to the $c_{cls}$, which brings improvement in recognition accuracy. In this case, $\mathcal L_{oti}$ and $\mathcal L_{match}$ mutually promote the training of the entire model.
\item {} $\theta_1+\theta_2<90^\circ$: As the decrease in $\theta_1+\theta_2$, $cos(\theta_1+\theta_2-90^\circ)$ decreases, causing $v_{otf}$ to move further away from the $v_{g2}$. The angle $\theta_3$ is larger, and $v_{af\_otf}$ is further away from $c_{cls}$. However, $\mathcal L_{oti}$ encourages $v_{af\_otf}$ to close $c_{cls}$. Therefore, $\mathcal L_{oti}$ and $\mathcal L_{match}$ constrain each other during training. 
\end{itemize}

\textbf{Inference.} 
Since unseen categories have not appeared during the training process, to achieve better ZSVR performances of the model, we use the refined spatial-temporal feature $v_{af\_otf}$ of video for classification. For a video from unseen categories, we extract its feature $v_{af\_otf}$ and semantic features of all unseen categories through orthogonal temporal interpolation module and the category branch, respectively. We then calculate the cosine similarity scores between the feature $v_{af\_otf}$ and semantic features of all unseen categories. Finally, we select the category with the highest similarity score as the predicted category of this video.

\section{Experiments}

\begin{table*}
\caption{ZSVR accuracies(\%) on UCF101 and HMDB51 with different protocols. ``Ob'' stands for Object scores.}
\label{tab:hmdb_ucf_two_split}
     \centering
    \begin{tabular}{ccclllllll}
    
    \hline
        Method & Video Feature & UCF101 & ~ & HMDB51 & ~ \\ 
        ~ & ~ & EP2 & EP1 & EP2 & EP1 \\\hline
        TS-GCN\cite{ts-gcn} & Ob(GoogleNet) & 36.1±4.8 & - & 23.2±3.0 & - \\ 
        TSRL\cite{tsrl} & Ob(CSWin) & 65.3±3.4 & - & 33.6±4.9 & - \\ \hline
        GA\cite{ga} & C3D & 17.3±1.1 & - & 19.3±2.119.3 & - \\ 
        CWEGAN\cite{cwegan} & I3D & 26.9±2.8 & - & 30.2±2.7 & - \\ 
        E2E\cite{e2e} & R(2+1)D & 48.0 & 37.6 & 32.7 / 26.9 & ~ \\ 
        ER\cite{ER} & TSM & 51.8±2.9 & - & 35.3±4.6 & - \\ 
        REST\cite{rest} & ResNet101 & 58.7±3.3 & 40.6 & 41.1±3.7 & 34.4 \\ \hline
        Text4Vis\cite{wu2022transferring} & VIT-L/14 & 85.8±3.3 & 79.6 & 58.1±5.7 & 49.8 \\ 
        BIKE\cite{bike} & VIT-L/14 & 86.4±3.2  & 80.6 & 59.4±3.6  & 49.8 \\ \hline
         ~& VIT-B/32 & 85.5±3.0 & 79.0 & 60.1±3.4 & 49.9 \\ 
        \textbf{OTI} & VIT-B/16 & 89.2±2.1 & 83.7 & 60.9±3.9 & 51.0 \\ 
        ~ & VIT-L/14 & \textbf{92.8±2.5} & \textbf{88.3} & \textbf{64.0±6.1} & \textbf{55.8} \\ \hline
    \end{tabular}
\end{table*}
\subsection{Experimental Setup}
\subsubsection{\textbf{Datasets}}
We conduct experiments on Kinetics-400~\cite{k400}, UCF101~\cite{UCF101}, HMDB51~\cite{HMDB51} and Kinetics-600~\cite{k600}. We train our model on videos from Kinetics-400, whose categories are regarded as the seen categories. We evaluate our model on videos from the remaining three unseen-category datasets. Specifically, Kinetics-400 consists of 400 human action categories, with at least 400 videos for each category. Kinetics-600 is an expansion of the Kinetics-400, including 220 new categories, and we evaluate our model on the new categories. UCF101 contains 13,320 videos divided into 101 categories, while HMDB51 consists of 6,766 videos in 51 categories.

\subsubsection{\textbf{Evaluation Protocols}}

\textbf{UCF101 and HMDB51}: According to \cite{e2e,x-clip}, there are three protocols for zero-shot evaluation:
\begin{itemize}
\item {} \textbf{Evaluation Protocol 1 (EP1)}: We test our model with videos from full categories. Therefore, more reliable evaluation can be conducted.
\item {} \textbf{Evaluation Protocol 2 (EP2)}: We randomly select half of the categories from all categories as the evaluation data. The final accuracy is calculated as the average result of the 10 evaluations. 
\item {} \textbf{Evaluation Protocol 3 (EP3)}: We test our model using three official splits and average the accuracies of the three splits for a fair comparison with the X-CLIP~\cite{x-clip} and ViFiCLIP~\cite{hanoonavificlip}.
\end{itemize} 

\textbf{Kinetics-600}: To maintain consistency, the same split from ~\cite{ER,bike,wu2022transferring} will be adopted. For this evaluation split, 160 categories will be selected from the 220 expanded categories, and the evaluation will be repeated three times to ensure reliability. We record the average accuracy of these three repetitions.

\subsubsection{\textbf{Implementation Details}}
For our video encoder, we utilize three different backbone architectures from CLIP~\cite{clip}: VIT-B/32, VIT-B/16 and VIT-L/14. Our model generates 49 patches per image with VIT-B/32, 196 patches with VIT-B/16, and 256 patches with VIT-L/14. We train a one-layer transformer to model temporal relationships. During the training phase, we randomly sample $T=8$ video frames along the temporal dimension for each video (experiments with uniform sampling shown in the Appendix~\ref{uniform_sample}). To optimize the model parameters, we use AdamW optimizer~\cite{loshchilov2017decoupled}. The VIT-B/32 based model and VIT-B/16 based model are trained with a learning rate of $5\times 10^{-5}$, while the VIT-L/14 based model is trained with a learning rate of $1\times 10^{-5}$. 
Automatic mixed precision (AMP)~\cite{lam2013automatically} is adopted to achieve improved performance and faster training process. The VIT-B/32 based model is trained for 5 epochs, while the VIT-B/16 based model and VIT-L/14 based model are trained for 10 epochs. The interpolation weight $\lambda$ in Eqn.~(\ref{eq:newf}) is set to 1 for videos from seen categories. 
The loss weights $\alpha$, $\beta$ and $\gamma$ in Eqn.~(\ref{eq:all_loss}) are all set to 1. During the inference phase, we sample multiple clips per video with several spatial crops according to~\cite{feichtenhofer2019slowfast,i3d} to achieve higher ZSVR accuracies. For HMDB51, we use two clips with three crops, while for UCF101 and Kinetics-600, we use three clips with three crops.

\subsection{Main Results}
\label{sec:main}

We compare the proposed OTI with some challenging zero-shot video action recognition methods. 
We categorize the previous state-of-the-art methods into three groups based on how the video features are obtained.
(1) Methods that mainly use indirect way to represent the video~\cite{ts-gcn,tsrl}; (2) Methods that use spatial-temporal convolution models to obtain video feature~\cite{ga,cwegan,e2e,ER,rest}; (3) CLIP-based methods that adapt the image-based CLIP to the video domain~\cite{wu2022transferring,bike,hanoonavificlip}. The ZSVR accuracies on UCF101 and HMDB51 under \textit{Evaluation Protocol 1} and \textit{Evaluation Protocol 2} are shown in Table~\ref{tab:hmdb_ucf_two_split}. 
Compared with the methods from the first two groups, CLIP-based methods, including Text4Vis~\cite{wu2022transferring} and BIKE~\cite{bike}, achieve significantly better performances.

When using the \textit{Evaluation Protocol 1} for evaluation, our model with the backbone VIT-L/14 outperforms the previous best competitor by {7.7\%} on UCF101 and {6.0\%} on HMDB51. It is worth noting that even though VIT-B/32 has fewer parameters than VIT-L/14, our model with the backbone VIT-B/32 still achieves competitive ZSVR accuracies compared with previous methods that use VIT-L/14 as the backbone.
Compared with the previous state-of-the-art competitor, the proposed OTI with the backbone VIT-B/16 achieves improvements of {1.2\%} on HMDB51 and {3.1\%} on UCF101.
When using \textit{Evaluation Protocol 2} for evaluation, the proposed OTI with the backbone VIT-L/14 achieves new state-of-the-art accuracies, reaching 92.8\% on UCF101 and 64.0\% on HMDB51.

\begin{table}[h]
\caption{ZSVR accuracies(\%) on UCF101 and HMDB51 with Evaluation Protocol 3.}
\label{tab:hmdb_ucf_three_split}
    \centering
    \begin{tabular}{cccc}
    \hline
        Method  & Encoder & UCF101 & HMDB51 \\ \hline
        X-CLIP\cite{x-clip} & VIT-B/16 & 72.0±2.3 & 44.6±5.2 \\ 
        ViFi-CLIP\cite{hanoonavificlip} & VIT-B/16 & 76.8±0.7 & 51.3±0.6 \\ \hline
         ~& VIT-B/32 & 78.6±1.6 & 52.3±0.6 \\ 
        \textbf{OTI} & VIT-B/16 & {83.3±0.3} & {54.2±1.3} \\ 
        ~ & VIT-L/14 &\textbf{88.1±1.0}  & \textbf{59.3±1.7} \\ \hline
    \end{tabular}
\end{table}
 
Table~\ref{tab:hmdb_ucf_three_split} shows the ZSVR accuracies with \textit{Evaluation Protocol 3}. The proposed OTI with the backbone VIT-B/16 outperforms X-CLIP by {11.3\%} and VIFI-CLIP by {6.5\%} on UCF101, and outperforms X-CLIP by {9.6\%} and VIFI-CLIP by {2.9\%} on HMDB51. Our method with the backbone VIT-B/32 has already surpassed the best previous method. Our method with the backbone VIT-L/14 achieves the best ZSVR accuracies, reaching {88.1\%} on UCF101 and {59.3\%} on HMDB51.
\begin{table}
    \caption{ZSVR accuracies(\%) on Kinetics-600.}
    \label{tab:k600}
    \centering
    \begin{tabular}{ccc}
    \hline
        Method & Encoder & Kinetics-600 \\ \hline
        ER\cite{ER} & TSM & 42.1±1.4 \\ \hline
        X-CLIP\cite{x-clip} & VIT-B/16 & 65.2±0.4 \\ 
        Text4Vis\cite{wu2022transferring}  & VIT-L/14 & 68.9±1.0 \\ 
        BIKE\cite{bike} & VIT-L/14 & 67.0±1.3 \\ \hline
        ~\ & VIT-B/32 & 64.5±1.1 \\ 
        \textbf{OTI} & VIT-B/16 & {66.9±1.0} \\ 
        ~ & VIT-L/14 & \textbf{70.6±0.5} \\  \hline
    \end{tabular}

\end{table}

Table~\ref{tab:k600} presents the ZSVR accuracies on Kinetics-600. Using the same backbone VIT-B/16, our method outperforms X-CLIP by {1.7\%}. Moreover, our method with the backbone VIT-L/14 outperforms Text4Vis by {1.7\%} and BIKE by {3.6\%}.
In summary, the proposed OTI achieves new state-of-the-art performances on UCF101, HMDB51, and Kinetics-600 datasets. These results validate the effectiveness of orthogonal temporal interpolation and matching loss.

\subsection{Ablation Studies}
\label{sec:abs}

In this section, we conduct additional ablation experiments to showcase the effectiveness of the proposed techniques. VIT-B/32 is adopted as the visual encoder in our method. We report the ZSVR accuracies on UCF101 and HMDB51 datasets with \textit{Evaluation Protocol 1}. The ablation study about the influence of residual connection is shown in Appendix~\ref{supp:resi_c}.

\begin{figure*}[htbp]
	\centering
        \includegraphics[width=0.96\textwidth]{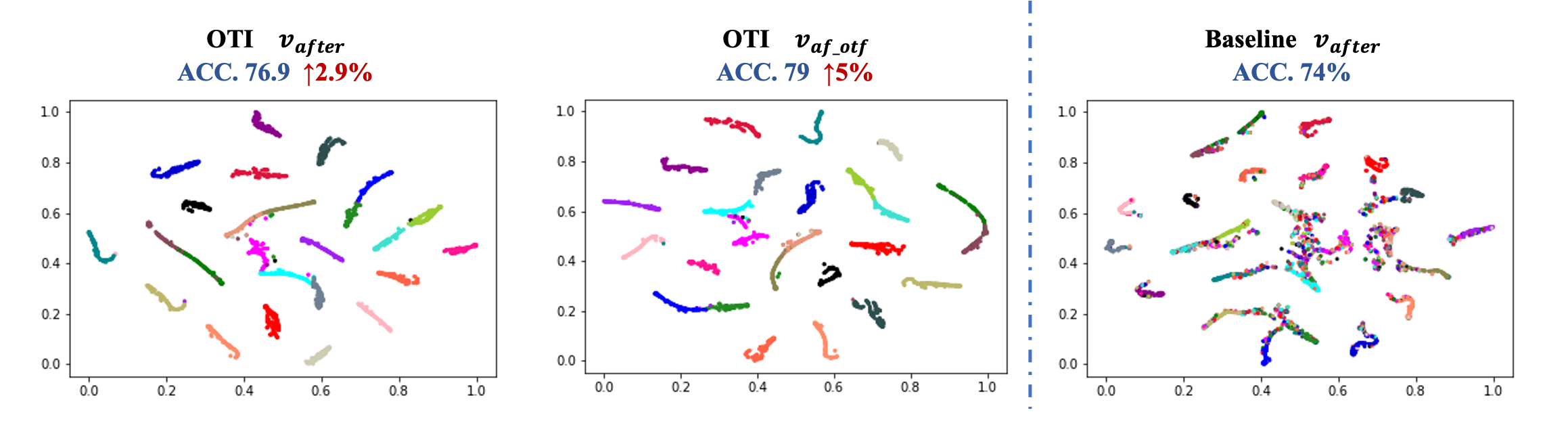}
	\caption{Using T-SNE to visualize the predictions of various trained models on UCF101. }
	\label{t-sne}
\end{figure*}

\subsubsection{\textbf{The Effects of $\mathcal L_{oti}$ and $\mathcal L_{match}$.}} 
\label{loss_co}

In order to verify the validity of $\mathcal L_{oti}$ and $\mathcal L_{match}$, we construct three different variants. The first variant is the baseline model trained with loss $\mathcal L_{cls}$. The second variant is the OTI model trained with loss $\mathcal L_{oti}$ and $\mathcal L_{cls}$. The third variant is the OTI model trained with additional loss $\mathcal L_{match}$, $\mathcal L_{oti}$ and $\mathcal L_{cls}$. Table~\ref{tab:loss} shows the ZSVR accuracies of the three different variants on UCF101 and HMDB51. We also provide visualizations of ZSVR accuracies of three different variants, as well as the trends of ZSVR accuracies of the third variant with different backbones. The results are presented in the Appendix~\ref{supp:loss}.

\begin{table}
\caption{the ZSVR accuracies (\%) of the three different variants on UCF101 and HMDB51.}

\label{tab:loss}
     \centering
    \begin{tabular}{ccccc}
    \hline
       ~& Loss  & $v_{before}$  & $v_{after}$ & $v_{af\_otf}$ \\ \hline
       HMDB51 & {$\mathcal L_{cls}$}  & 48.0 & 44.7 & 47.1\\ 
         & {$\mathcal L_{cls}$+$\mathcal L_{oti}$}  & {46.2} & 43.7 & {48.8}  \\ 
         & {$\mathcal L_{cls}$+$\mathcal L_{oti}$+$\mathcal L_{match}$}  & {46.57} & {46.61} & {49.9}  \\\hline
        UCF101 & {$\mathcal L_{cls}$}  & 77.8 & 74 & 78.4\\ 
         & {$\mathcal L_{cls}$+$\mathcal L_{oti}$}  & {75.9} & 76.2 & {78.8}  \\ 
         & {$\mathcal L_{cls}$+$\mathcal L_{oti}$+$\mathcal L_{match}$}  & {75.7} & {76.9} & {79.0}  \\
         \hline
    \end{tabular}
\end{table}

For the first variant, the model that uses the spatial feature $v_{before}$ outperforms the model using the spatial-temporal feature $v_{after}$ by 3.3\% and 3.8\% in accuracy on HMDB51 and UCF101, respectively. The ZSVR accuracy of the first variant using the feature $v_{af\_otf}$ is 0.9\% lower than the one of the first variant using feature $v_{before}$ on HMDB51, but it improves by 0.6\% on UCF101. 

We observe that the second variant using the feature $v_{af\_otf}$ improves by 2.6\% and 2.9\% on HMDB51 and UCF101, compared with the second variant using feature $v_{before}$. Compared with the first variant, the accuracy gaps between the model using the feature $v_{af\_otf}$ and the one using feature $v_{before}$ have increased by {3.5\%} ($48.8\%-46.2\%$ vs $47.1\%- 48.0\%$) on HMDB51 and 2.3\% ($78.8\%-75.9\%$ vs $78.4\%- 77.8\%$) on UCF101. The additional loss $\mathcal L_{oti}$ helps the model learn more effective spatial-temporal feature for video from unseen categories, which improves the transferability of the model.

With additional loss $\mathcal L_{match}$, the third variant using the feature $v_{af\_otf}$ achieves improvements of {1.1\%} and {0.2\%} on HMDB51 and UCF101, compared with of the second variant. The improved results suggest that our model learns a better refined spatial-temporal feature for video. Meanwhile, we can clearly observe that the model using the feature $v_{after}$ performs better than the one using the feature $v_{before}$ for both HMDB51 and UCF101.
The comparative results indicate that the trained temporal learning module can effectively learn temporal relationships for video from unseen categories. Therefore, the addition of $\mathcal L_{match}$ can encourage the model to learn better orthogonal temporal feature for video. Based on the analysis of the ZSVR results on HMDB51 and UCF101, 
we can conclude that joint training with  $\mathcal L_{cls}$, $\mathcal L_{oti}$ and $\mathcal L_{match}$ is meaningful.
\begin{table}
	\caption{The ZSVR accuracies (\%) of models with different number of layers in the temporal learning module on UCF101 and HMDB51.}
	\centering
	\begin{tabular}{lllll}
		\hline
		dataset & layers & $v_{before}$ & $v_{after}$ & $v_{af\_otf}$ \\ \hline
		HMDB51 & 1 & 46.57 & 46.61 & 49.9 \\ 
		HMDB51 & 2 & 45.7 & 42.3 & 48.6 \\
		UCF101 & 1 & 75.7 & 76.9 & 79.0 \\ 
		UCF101 & 2 & 75.1 & 75.4 & 78.5 \\ \hline
	\end{tabular}
	\label{tab:number_layer}
\end{table}
\subsubsection{\textbf{Different the Number of Layers for Temporal Learning.} }
\label{supp:nol}

To explore the impacts of different number of layers in the temporal learning module on ZSVR performance, we modify the 1-layer transformer to 2-layer transformer. For this experiment, we use VIT-B/32 as the visual encoder.
The ZSVR accuracies are shown in the Table~\ref{tab:number_layer}. We observe that the ZSVR accuracies of the model using the feature $v_{af\_otf}$ decrease by 0.5\% on UCF101 and by 1.3\% on HMDB51 when we increase the number of layers. Therefore, increasing the number of layers in the temporal learning module do not improve the model's zero-shot recognition ability.

\subsection{Visualization}

We use T-SNE~\cite{van2008visualizing} to illustrate the advantages of the proposed OTI intuitively. We select 25 categories of videos from UCF101, and obtain each video's prediction score. Then we project the prediction scores into 2 dimensional features using T-SNE. The visualization results are shown in Figure~\ref{t-sne}. We can see that using $v_{af\_otf}$, the videos within the same category are more concentrated. The confusion between different categories has decreased. This indicates that the proposed OTI learns more appropriate temporal information for videos from unseen categories and has a stronger transferability.
\section{Conclusion}

We observe an abnormal phenomenon in CLIP-based ZSVR methods for the first time, where the model that uses spatial-temporal feature performs much worse than the model that removes the temporal learning module and uses only spatial feature. We conjecture that improper temporal modeling on video disrupts the spatial feature of the video. To verify our conjecture, we propose feature factorization and observe that appropriate fusion of orthogonal temporal feature and spatial feature is effective in enhancing the feature representation ability in the ZSVR task. Based on these observations, we propose an orthogonal temporal interpolation module that learns better refined spatial-temporal feature. We also introduce a matching loss to improve the quality of the temporal feature. With the proposed techniques, we propose OTI for ZSVR, which achieves new state-of-the-art performances on three widely used ZSVR benchmarks.

\begin{acks}
This work was supported in part by the National Key R\&D
Program of China under Grant 2018AAA0102000, in part
by National Natural Science Foundation of China: 62022083 and 62236008.
\end{acks}

\bibliographystyle{ACM-Reference-Format}
% \bibliography{sample-base}
\bibliography{sample}
\clearpage
\nobalance
\appendix

\section{Interpolation using Different feature}
\label{supp:otf-af}
\begin{figure}[!ht]

        \begin{minipage}[]{0.4\textwidth}
		% \centering       
		\includegraphics[width=\textwidth]{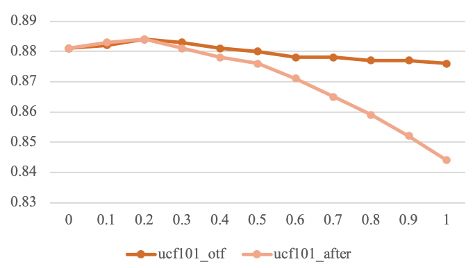}
            \label{fg:}
		\subcaption{ The comparison trends of ZSVR accuracies of pre-tained TextVis\cite{wu2022transferring} with the backbone VIT-L/14.}
	\end{minipage} 
        \quad
	\begin{minipage}[]{0.4\textwidth}
		\includegraphics[width=\textwidth]{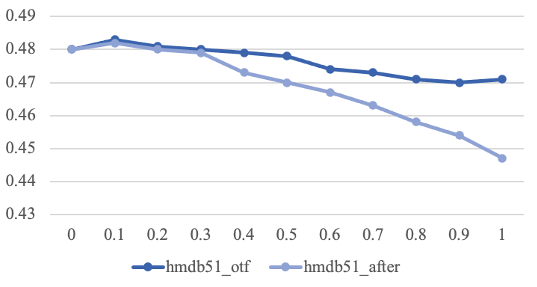}
            \label{fg:tm_text4vis}
		\subcaption{The comparison trends of ZSVR accuracies of our baseline model with the backbone VIT-B/32.}	
	\end{minipage} 
      
	\caption{The comparison trends of ZSVR accuracies of the model using different feature for interpolation.}
	\label{fig:otf-after}
\end{figure}

To verify the advantage of using orthogonal temporal feature $v_{otf}$, we replace feature $v_{otf}$ with feature $v_{after}$ (the difference between $v_{af\_res}$ and $v_{before}$) for interpolation fusion, and the trends of ZSVR accuracy are shown in Figure~\ref{fig:otf-after}. We find that when using feature $v_{after}$ for fusion, the ZSVR accuracy first increases and then sharply decreases for both UCF101 and HMDB51. And the decreasing trend is much steeper compared with the decreasing trend of using the feature $v_{otf}$ for fusion.
This result indicates that feature $v_{after}$ causes greater damage to the spatial feature $v_{before}$. On the other hand, using orthogonal temporal feature $v_{otf}$ performs much more stably.

\section{Different methods of sampling video frames}
\label{uniform_sample}
\begin{table}[!ht]
    \centering
 
    \caption{The ZSVR accuracies (\%) of the proposed OTI using random and uniform sampling on HMDB51 and UCF101.}
    
    \begin{tabular}{lllll}
    \hline
         
         ~ & backbone & Text4vis & OTI\_uniform & OTI\_random \\ \hline
        HMDB51 &VIT-L/14 & 49.8 & 56.5 & 55.8 \\ 
        HMDB51 & VIT-B/32 & & 49.9 & 49.5 \\
        UCF101 &VIT-L/14 & 79.6 & 88.7 & 88.3 \\ 
        UCF101 & VIT-B/32 & &79.0 & 78.8 \\ \hline
        
    \end{tabular}
    \label{uni_rand}
\end{table}

To investigate the impact of different video frame sampling strategies on ZSVR performance, we also conduct experiment using uniform sampling, which is adopted in Text4Vis~\cite{wu2022transferring}. The comparison results are shown in Table~\ref{uni_rand}, which demonstrate that the proposed OTI still maintains strong ZSVR ability when using the same uniform sampling as Text4vis. Additionally, the results also indicate that our model is stable across various sampling methods. 

\section{Supplement of Ablation Studies}
\label{supp:abla}

\subsection{More Illustration of Effects of $\mathcal L_{oti}$ and $\mathcal L_{match}$}
\label{supp:loss}
\begin{figure}[htbp]
        
	\centering
    \includegraphics[width=0.45\textwidth]{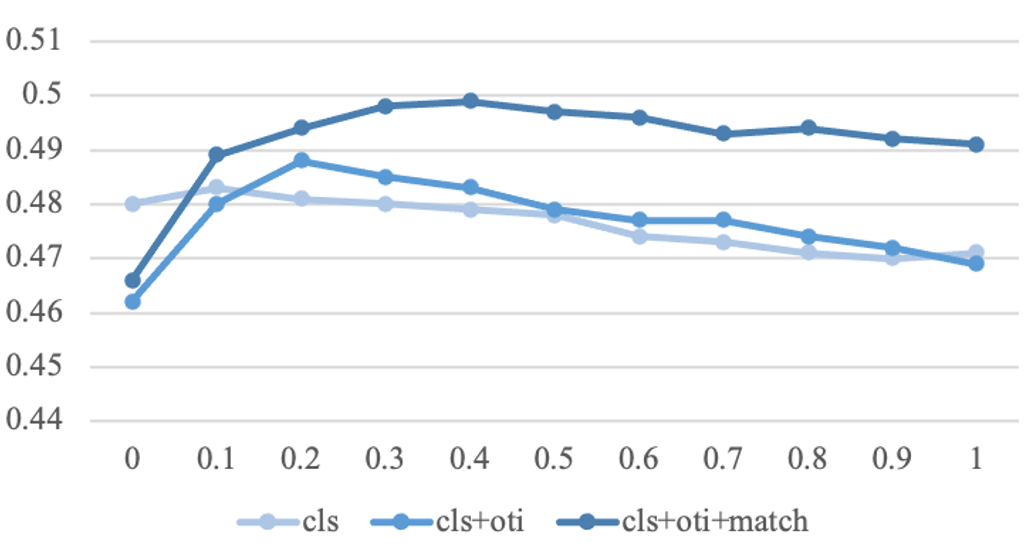}
	\caption{The comparison trends of ZSVR accuracies of three different variants on HMDB51.}
	\label{fig:lossess} 
\end{figure}
To visually demonstrate the impact of $\mathcal L_{oti}$ and $\mathcal L_{match}$, we use three different variants for the ZSVR task. These variants include a model trained with loss $\mathcal L_{cls}$, a model trained with loss $\mathcal L_{cls}$ and $\mathcal L_{oti}$, a model trained with loss $\mathcal L_{cls}$, $\mathcal L_{oti}$ and $\mathcal L_{match}$. The third variant is the proposed OTI.
The trends of ZSVR accuracies are shown in Figure~\ref{fig:lossess}. 

For the first variant, the ZSVR accuracy of the model using the feature $v_{af\_otf}$  is lower than the one of the model using the feature $v_{before}$ when $\lambda>0.3$. This indicates that the temporal feature learned by the current temporal learning module have a negative impact on the spatial feature.
For the second variant, as $\lambda$ increases, the overall ZSVR accuracy first increases and then decreases, but the ZSVR accuracy of the model using the feature $v_{af\_otf}$ is always higher than the one of the model using the feature $v_{before}$. These results indicate that the additional $\mathcal L_{oti}$ helps the model learn more effective spatial-temporal feature for video from unseen categories.
For the third variant, as $\lambda$ increases, the ZSVR accuracy first significantly increases and then gets stabilized. The addition of $\mathcal L_{match}$ enables the model to learn better orthogonal temporal feature. In summary, training the model jointly with $\mathcal L_{cls}$, $\mathcal L_{oti}$ and $\mathcal L_{match}$ is effective.

\begin{figure}[htbp]
        
	\centering
    \includegraphics[width=0.47\textwidth]{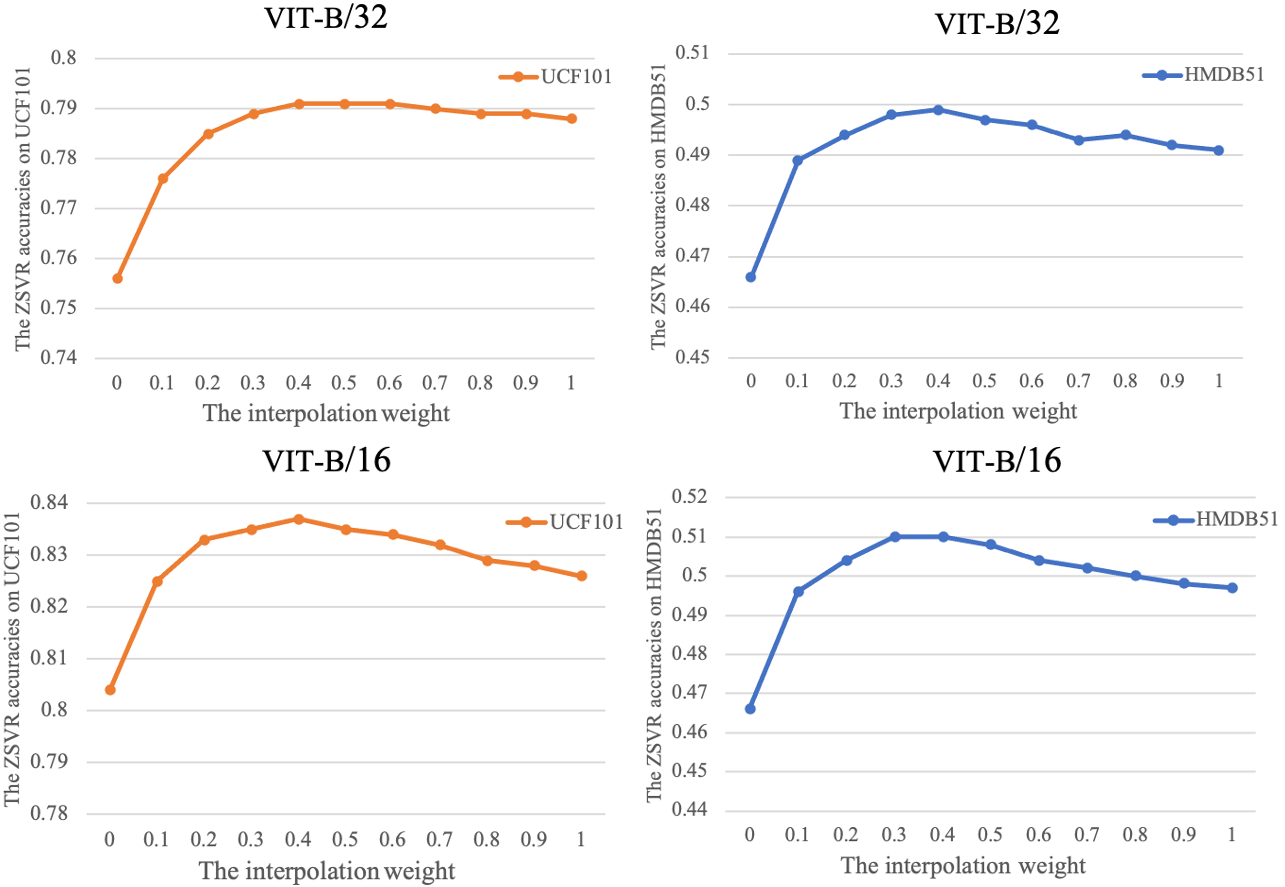}
   
	\caption{The trends of ZSVR accuracies of models with different backbones on HMDB51 and UCF101.}
	\label{ratio}

\end{figure}

At the same time, in order to visually demonstrate the effectiveness of the proposed OTI over different backbones, we also presents the ZSVR accuracies of OTI with the backbone VIT-B/32 and VIT-B/16. The refined spatial-temporal $v_{af\_otf}$ is used for ZSVR. The experimental results are depicted in Figure~\ref{ratio}. It can be observed that, for both backbones, the ZSVR accuracy exhibits a similar trend on UCF101 and HMDB51. As the interpolation weight $\lambda$ increases, the ZSVR accuracy shows a rapid improvement initially, followed by stabilization with minor fluctuations.
Each benchmark has its optimal $\lambda$ for achieving the best ZSVR accuracy, indicating that appropriate temporal modeling for video is necessary for the ZSVR task.

\subsection{The Influence of Residual Connection}
\label{supp:resi_c}

\begin{table}[!h]
\caption{The ZSVR accuracies (\%) of models with or without residual connection operation on UCF101 and HMDB51.}

\label{tab:res}
     \centering
    \begin{tabular}{ccccc}
    \hline
       ~& Residual & $v_{before}$  & $v_{after}$ & $v_{af\_otf}$ \\ \hline
       HMDB51 & w/o   & 46.57 & 46.61 & 49.9\\ 
         & w/  & 45.2 & 39.9 & 48.7  \\ \hline
        UCF101  & w/o  & 75.7 & 76.9 & 79.0\\
        & w/  & 75.6 & 72.7 & 78.2 \\
         \hline
    \end{tabular}
\end{table}

To explore the influence of residual connection, we also train a model with residual connection and conduct evaluation experiments on UCF101 and HMDB51. The ZSVR accuracies are presented in Table~\ref{tab:res}. 
Compared with the model without residual connection, the ZSVR accuracies of the model with residual connection using feature $v_{af\_otf}$ drop by 1.2\% and 0.8\% on HMDB51 and ucf101. The ZSVR accuracies of the model using feature $v_{before}$ are significantly higher than those of the model using feature $v_{after}$. Therefore, we can conclude that for the ZSVR task, when the spatial-temporal feature of the video contains too much spatial feature, the optimization of the parameters from the temporal learning module may be constrained during the training process.

\end{document}